# VS-TransGRU: A Novel Transformer-GRU-based Framework Enhanced by Visual-Semantic Fusion for Egocentric Action Anticipation

Congqi Cao
congqi.cao@nwpu.edu.cn
Northwestern Polytechnical
University
China

Ze Sun
ze.sun@mail.nwpu.edu.cn
Northwestern Polytechnical
University
China

Qinyi Lv
Northwestern Polytechnical
University
China

Lingtong Min
Northwestern Polytechnical
University
China

Yanning Zhang
Northwestern Polytechnical
University
China

## ABSTRACT

Egocentric action anticipation is a challenging task that aims to make advanced predictions of future actions from current and historical observations in the first-person view. Most existing methods focus on improving the model architecture and loss function based on the visual input and recurrent neural network to boost the anticipation performance. However, these methods, which merely consider visual information and rely on a single network architecture, gradually reach a performance plateau. In order to fully understand what has been observed and capture the dependencies between current observations and future actions well enough, we propose a novel visual-semantic fusion enhanced and Transformer-GRU-based action anticipation framework in this paper. Firstly, high-level semantic information is introduced to improve the performance of action anticipation for the first time. We propose to use the semantic features generated based on the class labels or directly from the visual observations to augment the original visual features. Secondly, an effective visual-semantic fusion module is proposed to make up for the semantic gap and fully utilize the complementarity of different modalities. Thirdly, to take advantage of both the parallel and autoregressive models, we design a Transformer-based encoder for long-term sequential modeling and a GRU-based decoder for flexible iteration decoding. Extensive experiments on two large-scale first-person view datasets, i.e., EPIC-Kitchens and EGTEA Gaze+, validate the effectiveness of our proposed method, which achieves new state-of-the-art performance, outperforming previous approaches by a large margin.

## KEYWORDS

Egocentric Action Anticipation, Semantic, Visual-Semantic Fusion, Transformer, GRU

## 1 INTRODUCTION

Egocentric (first-person) action anticipation is an essential component of artificial intelligence and computer vision, with a wide range of application values. For example, in autonomous driving [30, 33], vehicles need to determine whether pedestrians will cross the intersection based on their current and past behaviors. In human-computer interaction systems [24, 35], if machines can anticipate people's subsequent actions and provide feedback accordingly, it will bring a higher-quality experience for users.

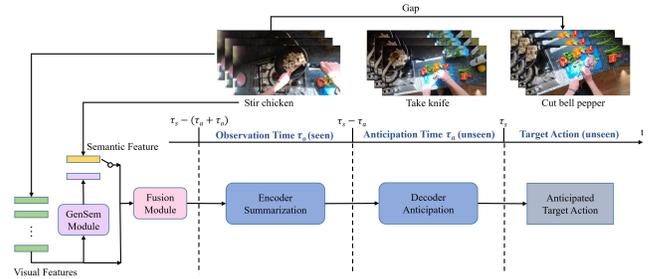

Figure 1: Illustration of action anticipation task and our solution. Observing action starting at time $\tau_s-(\tau_a+\tau_o)$ and ending at time $\tau_s - \tau_a$, the task objective is to anticipate future action at time $\tau_s$ after an interval of $\tau_a$, where $\tau_s$ represents the starting moment of the target action, $\tau_a$ represents the anticipation time, and $\tau_o$ represents the observation length. Due to the gap between the observed and future actions, anticipating the future is more challenging than recognizing the present. We propose to generate semantic features based on category labels or visual cues, then fuse them with visual features and input the fused features into an encoder-decoder model to anticipate future actions.

Formally, action anticipation is anticipating future actions based on current and historical observation data. As shown in Figure 1, the future action "Cut bell pepper" is anticipated by observing the already occurred action "Stir chicken." Besides the visible observation, the content of the anticipation stage and the target moment is invisible to the model.

Egocentric action anticipation is highly challenging. Although both involve modeling observation data, action recognition aims to identify the current action category, while egocentric action anticipation seeks to anticipate future action. Future actions have apparent visual differences and logical connections with the current observations, making it difficult to fully understand the observed data and capture the relationship between it and future actions. The difficulty also results in generally low anticipation performance for mainstream methods that rely on visual information. Furthermore, the occurrence of future actions is influenced by various factors, leading to substantial uncertainty in future actions, which undoubtedly increases the difficulty of anticipation while also posing higher demands on the model's ability to mine the temporal relationships in the observations.

In egocentric action anticipation, classic methods are based on recurrent neural network (RNN) [9, 20]. To improve model performance further, numerous works have improved network structures and anticipation algorithms [15, 18, 27, 32, 38, 39, 45, 48]. However, current action anticipation methods have gradually reached a performance plateau. Analyzing these methods, most of them use RGB, optical flow, object or a combination of these modalities as input. In other words, they only focus on visual information in videos. We argue that relying solely on visual information cannot satisfy the need of action anticipation tasks, resulting in a performance bottleneck. The previous work [3] of third-person anticipation inputs the one-hot encoding of the ground truth labels of the observed action sequences into the network to test the performance ceiling on the scripted datasets. However, one-hot encoding has high sparsity, redundancy, and feature dimensions, and most importantly, lacks semantic correlation. Moreover, using only one-hot encoding ignores the rich visual context information in the videos, making it unsuitable for unscripted first-person action anticipation tasks. Subsequent outstanding work [19] on third-person anticipation has also shifted to using visual features as model inputs.

Appropriate inputs are critical to the model. However, few attempts are made on the input modality. Most researchers neglect the important high-level semantic information in videos, which can help the model better understand the current observations and capture the dependencies between the present and future, and instead focus solely on visual clues. In addition, to improve the anticipation accuracy of future actions, the methods proposed by researchers are becoming increasingly complicated and even cumbersome. In order to solve the challenges faced by egocentric action anticipation task, break through the performance bottleneck of existing work using visual information, and make the model structure as concise as possible, we propose a novel Transformer-GRU-based framework enhanced by visual-semantic fusion for egocentric action anticipation (VS-TransGRU), as shown in Figure 3.

Specifically, we adopt an encoder-decoder structure, where the encoder summarizes the observations and the decoder anticipates the future based on the encoder's summary. Considering the significant achievements of Transformer [41] in natural language processing and computer vision in recent years, we utilize Transformer with its powerful temporal modeling capabilities as the encoder and employ GRU with its flexible iteration feature as the decoder to address different anticipation time issues. A large number of comparative experiments verify the effectiveness of the proposed model structure. Besides visual features, we propose a new semantic feature and design two ways to generate it. The first way is to pass the ground truth text label of the observed action into a pre-trained language model, and the output embedding of the language model is called the ground truth semantic feature. Since the semantic feature we propose is an entirely new modality of input that has not been used in other works, to demonstrate its effectiveness further and provide a fair comparison with state-of-the-art methods that only use visual information, we have designed the second way. The second way is to train a semantic generation module during training, allowing it to generate semantic features based on visual information. Semantic features can better represent the high-level semantic information of the observed actions, but there is still a gap between them and visual clues. Therefore, we propose several fusion methods to eliminate this gap, including learnable weighted sum, concatenation, multi-layer perceptron based, and attention-based techniques. Through experiments, we found that the learnable weighted sum fusion strategy is most conducive to fusing two types of features. The fused features are then fed into our proposed encoder-decoder model to anticipate future action. We generate anticipation loss at the target action at training time regardless of using ground truth or estimated semantic features. The difference is that, when using the estimated semantic features, there will also be classification loss for the current observed actions, in addition with cosine similarity and mean squared error loss for the estimated semantic features.

Experimental results show that our Transformer-GRU model has already surpassed most state-of-the-art methods when using only visual features. After introducing our proposed semantic features, the anticipation results outperform all existing methods. Particularly when using the ground truth semantic features, our framework achieves an improvement of up to 10.26% in Top-5 accuracy compared to the current best-performing method [32], which is a significant increase. Even using estimated semantic features, there is a performance improvement of up to 1.02% (the aforementioned data is based on Table 1 at 8 anticipation times).

In summary, our contributions are as follows: 1) To our best knowledge, we are the first to propose the use of semantic features to assist action anticipation, which is a new input modality different from visual features. Experimental results show that introducing semantic features can effectively improve anticipation performance. 2) We propose a simple and effective temporal reasoning architecture based on Transformer and GRU for egocentric action anticipation. Based on this model, we explore the best fusion strategy for semantic and visual features through extensive experiments. 3) With the proposed VS-TransGRU model, our method achieves state-of-the-art performance on large-scale first-person datasets such as EPIC-Kitchens and EGTEA Gaze+, significantly surpassing previous state-of-the-art methods.

## 2 RELATED WORK

Action anticipation can be divided into long-term and short-term anticipation, depending on the length of the anticipation time $\tau_a$.
**Long-term Action Anticipation.** Long-term anticipation methods are usually tested on scripted third-person datasets [1–3, 16, 19, 23, 28], such as 50 Salads [37] and Breakfast [25]. Depending on the dataset, the anticipation time can be as long as several minutes. Abu et al. [1, 3] first proposed a method using RNN and CNN to simultaneously anticipate future action categories and durations, and later used two RNNs to model future action categories and durations separately. Loh et al. [28] used a multi-headed attention-based variational RNN and Gaussian log-likelihood maximization to anticipate the categories and durations of future actions. Ke et al. [23] used action label sequences and anticipation time as model inputs and employed temporal convolution, attention mechanism, and skip connections to anticipate the action category at a future moment directly. Gong et al. [19] used Transformer to model the dependencies between observations and future actions, added a classification loss for current observed content, and adopted parallel decoding to directly anticipate the categories and durations of future action sequences.

**Short-term Action Anticipation.** Short-term action anticipation methods are usually based on unscripted first-person large-scale datasets [13, 15, 18, 21, 27, 31, 32, 34, 36, 38, 39, 45–48], such as EPIC-Kitchens [10] and EGTEA Gaze+ [26], with anticipation time generally ranging from 0.25 seconds to 2 seconds. Furnari et al. [15] first used two LSTMs to encode and summarize inputs and make cyclic anticipations for the future. They employed a learnable attention mechanism to fuse anticipations from RGB, optical flow, and object modalities. Qi et al. [32] mitigated the error accumulation problem by using a contrastive loss and attention mechanism to revise intermediate features twice based on GRU and simultaneously anticipated action and verb/noun categories at the target moment, providing more supervision information. Liu et al. [27] added a classification loss for observed content, used an external memory bank to enhance network memory capacity, employed contrastive loss to correct intermediate anticipated results, and mapped the anticipated features at the target moment to make them as similar as possible to the ground truth feature sequences. Girdhar et al. [18] proposed the Anticipative Video Transformer (AVT), which used the Transformer encoder to process raw video frame sequences and the masked decoder to generate intermediate and target moment features and categories. Osman et al. [31] drew inspiration from action recognition and used slow and fast streams with different frame rates, providing some complementary information. Following previous work [15, 27, 32], our model uses an encoder-decoder structure. To take full advantage of the modeling characteristics of Transformer and RNN, we use them separately as encoder and decoder, rather than the conventional approach of using RNN as both encoder and decoder. In addition, previous methods focused only on visual information, making it increasingly difficult to improve anticipation performance. We propose a novel and practical semantic feature that can help the model better understand observed content and accurately capture future and current relationships. We also explore effective fusion methods between semantic and visual features. The proposed method significantly improves anticipation performance.

## 3 APPROACH
### 3.1 Task Formulation

The purpose of action anticipation is to anticipate possible actions after a specific time by observing a given video. As shown in Figure 1, given a video segment that starts at $\tau_s - (\tau_a + \tau_o)$ and ends at $\tau_s - \tau_a$, the task is to anticipate the action category after $\tau_a$. Here, $\tau_o$ represents the length of the observation, $\tau_a$ represents the anticipation time, i.e., the time interval between the observed video and the target action, and $\tau_s$ denotes the start time of the target action. It is important to note that the anticipation stage and target moments' content are not visible to the model.

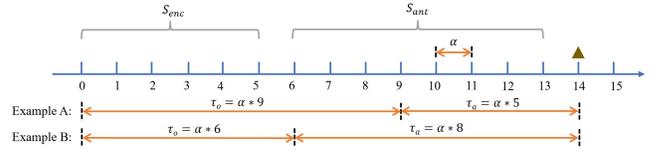

Figure 2: Variable input sequences and anticipation times. We take data from $S_{enc}$+$S_{ant}$ time-steps before the start of the target action. As commonly used, $S_{ant}$ is 8, representing the maximum anticipation steps. $S_{enc}$ is 6, representing the basic encoding steps. The interval between every two time-steps is 0.25 seconds. The range of anticipation steps is from 1 to 8, corresponding to 8 different input sequence lengths. "Example A" and "Example B" represent input sequence lengths of 9 and 6 time-steps, with anticipation steps of 5 and 8, respectively.

As shown in Figure 2, following previous work [15], assuming that the target action starts at time-step 14, we take $S_{enc}$+$S_{ant}$ ($S_{enc}$=6, $S_{ant}$=8) time-steps inputs before the start of the target action, with an interval of $\alpha$=0.25s between adjacent time-steps. The inputs are divided into 8 anticipation times: 2s, 1.75s, 1.5s, 1.25s, 1s, 0.75s, 0.5s, and 0.25s. Different anticipation times also correspond to different observation lengths. As shown in Figure 2, for "Example A" and "Example B", the observation lengths are 2.25s and 1.5s, while the anticipation times are 1.25s and 2s, respectively.

### 3.2 Framework Overview

The overview of our proposed framework is shown in Figure 3. We will introduce the three components that make up the overall framework in the following.

**Visual-Semantic Preprocessing.** This part mainly converts the original video and category labels into semantic-enhanced visual features. The original frame sequence or video clip sequence $\{V_1, V_2, \cdots, V_{t_o}\}$ is first passed through a feature extractor $\varphi$ to obtain visual features. The feature extractor can be action recognition networks such as TSN [43], I3D [8], irCSN152 [40], ViViT [4], as shown below:

$$f_i = \varphi(V_i) \tag{1}$$

where $f_i \in \mathbb{R}^{1 \times d_v}$, $d_v$ represents the dimension of visual features, and $i = 1, 2, \cdots, t_o$. Semantic features represent the high-level semantic information of the current observation and have strong semantic relevance, which can effectively assist the model in anticipation. They can be obtained by inputting the ground truth category labels into a pre-trained language model, or estimated by our proposed model based on visual information. The two generation methods of semantic features will be specifically introduced in Section 3.3.

After obtaining the visual and semantic features, the fusion module $\emptyset$ combines the two, outputting semantic-enhanced visual features. The strategies adopted by the fusion module will be introduced in Section 3.4. Let $F = \{f_1, f_2, \cdots, f_{t_o}\} \in \mathbb{R}^{t_o \times d_v}$, the fusion process is as follows:

$$X = \emptyset(s, F) \tag{2}$$

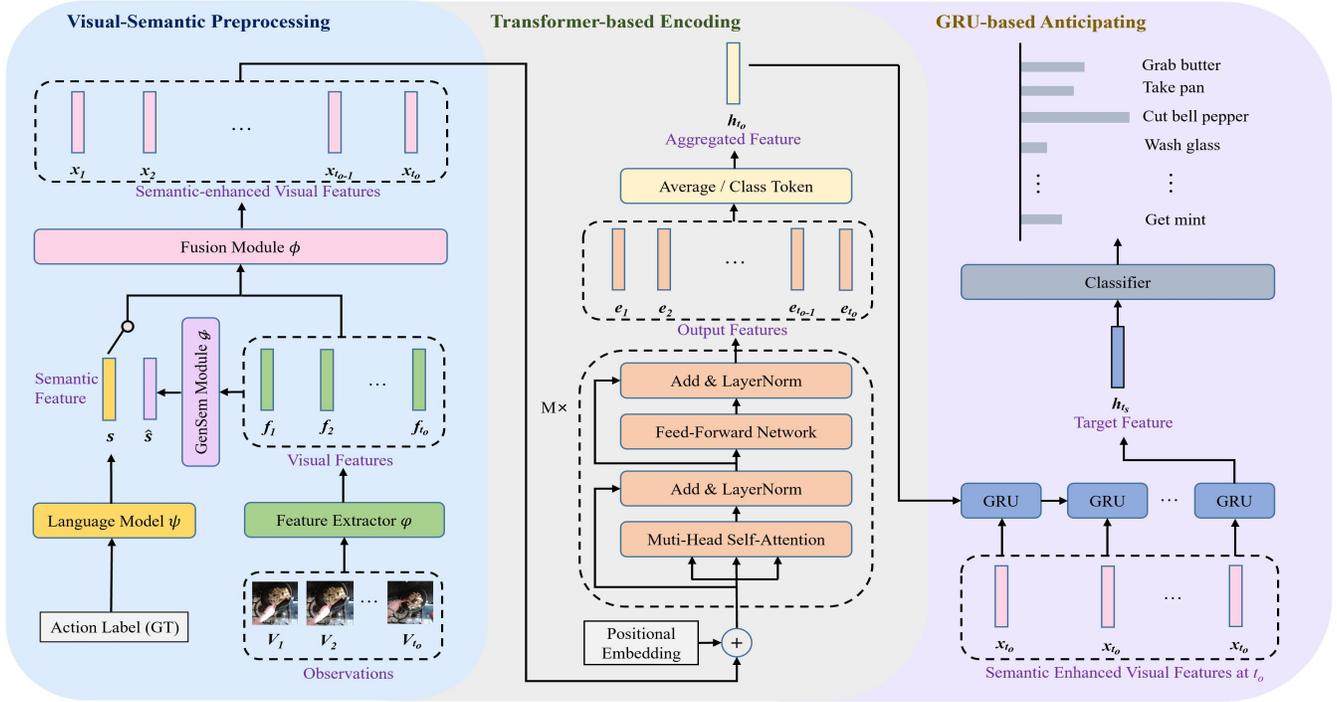

Figure 3: Overview of the proposed VS-TransGRU framework. VS-TransGRU is divided into three parts. The first part involves extracting visual features, generating semantic features, and fusing the two types of features. The second part encodes the fused features using a Transformer, obtaining summarized features. The third part generates anticipation results for future actions by iteratively using a GRU based on the Transformer's summarized features.

where $X = \{x_1, x_2, \cdots, x_{t_o-1}, x_{t_o}\} \in \mathbb{R}^{t_o \times d_x}$, $d_x$ represents the dimension of the fused features, $s \in \mathbb{R}^{1 \times d_s}$ represents the semantic feature, and $d_s$ represents the dimension of semantic features.

**Transformer-based Encoding.** As described in the Introduction, we use Transformer to encode the observed data. Since the Transformer is not sensitive to the sequence, and the semantic-enhanced visual features are ordered, we need to add positional encoding to the feature sequence first. To handle variable-length input sequences, we adopt 2D fixed positional encoding, as shown below:

$$\text{PE}_{(pos,2i)} = \sin\left(\frac{pos}{10000^{\frac{2i}{d_x}}}\right) \quad (3)$$

$$\text{PE}_{(pos,2i+1)} = \cos\left(\frac{pos}{10000^{\frac{2i}{d_x}}}\right) \quad (4)$$

where $pos$ represents the sequence position, and $i$ represents the dimension position of the positional encoding vector. After adding the positional encoding, the sequence $X$ is passed into $M$ Transformer blocks. Each Transformer block consists of a multi-head self-attention mechanism, residual connections with layer normalization, and a feed-forward neural network. The multi-head self-attention mechanism introduces learnable parameters, allowing it to be continuously optimized during training so that different heads can learn different patterns. The propagation process of each Transformer block is as follows:

$$X'_m = \text{LN}\left(\text{MHSA}\left(X_{m-1}\right) + X_{m-1}\right) \quad (5)$$

$$X_m = \text{LN}(\text{FFN}(X'_m) + X'_m) \quad (6)$$

The Transformer does not change the dimension of the input feature sequence, meaning the output and input sequences maintain the same shape, as shown below:

$$E = \text{Transformer}(X) \quad (7)$$

where $E = \{e_1, e_2, \cdots, e_{t_o-1}, e_{t_o}\} \in \mathbb{R}^{t_o \times d_e}$, $d_e = d_x$, and $d_e$ represents the dimension of the Transformer output features. Therefore, to obtain the aggregated feature of the semantic-enhanced visual feature sequence, we need to take the mean of the Transformer's output or use the Class Token value as the global feature $h_{t_o}$.

**GRU-based Anticipation.** GRU can flexibly cope with variable anticipation times through different iterations, so we choose the GRU model for decoding. The semantic and visual summary feature $h_{t_o}$ is used as the initial hidden state of the GRU. Intuitively, the future action is more relevant to the content observed closest to it, so we use the semantic-enhanced visual feature $x_{t_o}$ at time $t_o$ as the input for the GRU. The number of iterations of the GRU is consistent with the anticipation time. Taking "Example A" in Figure 2 as an example, the GRU needs to iterate 5 times for anticipation. We call the hidden state of the last iteration of the GRU the anticipated target feature, which is input into the classifier to obtain the action

category scores at the target moment. The entire process is as follows:

$$h_0 = h_{t_o} \quad (8)$$
$$h_{i+1} = \text{GRU}(x_{t_o}, h_i) \quad (9)$$
$$y = \text{Classifier}(h_{t_s}) \quad (10)$$

where $y \in \mathbb{R}^{1 \times N}$ represents each category's anticipation scores, $N$ represents the number of action categories.

### 3.3 Semantic Feature Generation

**Ground Truth Semantic.** As described in Section 3.2, the text form of the ground truth category label of the current observed action is input into a pre-trained language model $\psi$, such as Phrase-Bert [44] or GPT-3 [6], to obtain the encoded feature. We call this feature the ground truth semantic feature. The encoding process is shown as follows:

$$s = \psi(label) \quad (11)$$

where $s \in \mathbb{R}^{1 \times d_s}$, $d_s$ represents the dimension of the semantic feature.

**Estimated Semantic.** We have designed a method to generate semantic features using visual information. The semantic features generated in this way are estimated semantic features. While anticipating future actions, [13, 18, 19, 27] use the current observed action category label as supervision information to generate classification loss, encouraging the model to model the observed data better. We adopt the same approach, generating classification loss for the current observation. What sets our work apart from theirs is that we further use the classification probability of the current observation category as a weight. This weight, combined with the matrix of ground truth semantic features corresponding to all categories, results in a weighted sum to obtain the generated semantic feature. Let $\omega \in \mathbb{R}^{1 \times N}$ represent the classification probability for the current observation, i.e., the output of the Softmax activation function. $S \in \mathbb{R}^{N \times d_s}$ represents the ground truth semantic feature matrix, and their weighted sum is shown as follows:

$$\hat{s} = \omega @ S \quad (12)$$

where @ represents matrix multiplication, and $\hat{s} \in \mathbb{R}^{1 \times d_s}$ represents the estimated semantic feature.

### 3.4 Visual-Semantic Fusion

There is a gap between semantic features and visual features, and semantic features are single features, while visual features are in the form of a sequence. To effectively integrate the two types of features, we adopt the following strategies:

**Concatenation.** We directly concatenate the semantic and visual features on the feature dimension, which is a simple fusion method and a common approach when combining different features.

**Learnable weighted sum.** When the dimensions of semantic and visual features are inconsistent, we use a linear layer to map the semantic features to the dimension of the visual features. Then, the two types of features are multiplied by their respective weights and summed to complete the fusion. Our experiments use learnable weights, allowing the model to learn the most suitable weights for different modalities during training.

**MLP fusion.** Based on the concatenation of the two features, we input the concatenated features into a multilayer perceptron (MLP) with two layers for fusion. The MLP is continuously optimized during training.

**Attention fusion.** Use the result of the mapped semantic features as the query $q \in \mathbb{R}^{1 \times d_v}$ and the visual feature sequence as the key and value $K, V \in \mathbb{R}^{t_o \times d_v}$ for attention calculation. The output of the attention is then concatenated with the semantic feature. The attention calculation process is as follows:

$$attn = \text{softmax}\left(\frac{qK^T}{\sqrt{d_v}}\right) V \quad (13)$$

where $attn \in \mathbb{R}^{1 \times d_v}$ represents the output of the attention. We propose these four fusion methods for integrating visual and semantic features.

### 3.5 Training Objective

We only generate anticipated action losses at the target time when using ground truth semantic features and visual features fusion. To further alleviate the problem of model overfitting, we use a cross-entropy loss function with label smoothing:

$$L_{tgt} = -(1-\theta) \sum_{i=1}^{N} y_i \log(\hat{y}_i) - \frac{\theta}{N} \sum_{i=1}^{N} \log(\hat{y}_i) \quad (14)$$

where $N$ represents the number of categories, $y_i$ represents the probability of the $i$-th category in the ground truth labels (0 or 1), and $\hat{y}_i$ represents the predicted probability of the $i$-th category by the model. $\theta$ is a hyperparameter with a value between 0 and 1, used to control the smoothing degree of the ground truth labels.

When using estimated semantic features, in addition to generating loss at the target time, we also use a cross-entropy loss function with label smoothing to calculate the classification loss of the current observed action. Furthermore, to make the estimated semantic features closer to the ground truth semantic features, we add two extra loss functions between them: cosine similarity and mean squared error (MSE) loss. These two loss functions focus on the differences in direction and magnitude between vectors, respectively, and have a certain complementarity, which is beneficial for optimizing the model. The cosine similarity and mean squared error losses are as follows:

$$L_{\cos} = 1 - \frac{\hat{s} \cdot s}{\|\hat{s}\| \cdot \|s\|} \quad (15)$$

$$L_{mse} = \sum_{i=1}^{d_s} (s_i - \hat{s}_i)^2 \quad (16)$$

where $\hat{s}$ represents the estimated semantic features and $s$ represents the ground truth semantic features. The total loss is as follows:

$$L = L_{tgt} + aL_{obs} + bL_{\cos} + cL_{mse} \quad (17)$$

where $L_{obs}$ represents the classification loss of the observed action, consistent with the loss function of the target action, $a, b, c$ are hyperparameters, representing the weights of different loss terms, respectively.

During inference, if using ground truth semantic features, the model needs to be fed with both ground truth semantic and visual features. If using estimated semantic features, only visual features are needed, and the semantic features are estimated by the model based on the visual information.

Table 1: Anticipation results of different methods on EPIC-Kitchens dataset. "GTS" means "Ground Truth Semantic", and "ES" means "Estimated Semantic".

| Methods | Top-5 Action Accuracy(%) @ Different Anticipation Times | | | | | | | | Top-5 Acc.(%) @ 1s | | | Mean Top-5 Rec.(%) @ 1s | | |
|---|---|---|---|---|---|---|---|---|---|---|---|---|---|---|
| | 2.00 | 1.75 | 1.50 | 1.25 | 1.00 | 0.75 | 0.50 | 0.25 | Verb | Noun | Action | Verb | Noun | Action |
| DMR [42] | / | / | / | / | 16.86 | / | / | / | 73.66 | 29.99 | 16.86 | 24.50 | 20.89 | 3.23 |
| ATSN [10] | / | / | / | / | 16.29 | / | / | / | 77.30 | 39.93 | 16.29 | 33.08 | 32.77 | 7.06 |
| VN-CE [10] | / | / | / | / | 17.31 | / | / | / | 77.67 | 39.50 | 17.31 | 34.05 | 34.50 | 7.73 |
| MCE [14] | / | / | / | / | 26.11 | / | / | / | 73.35 | 38.86 | 26.11 | 34.62 | 32.59 | 6.50 |
| SVM-TOP3 [5] | / | / | / | / | 25.42 | / | / | / | 72.70 | 28.41 | 25.42 | 41.90 | 34.69 | 5.32 |
| SVM-TOP5 [5] | / | / | / | / | 24.46 | / | / | / | 69.17 | 36.66 | 24.46 | 40.27 | 32.69 | 5.23 |
| VNMCE+T3 [14] | / | / | / | / | 25.95 | / | / | / | 74.05 | 39.18 | 25.95 | 40.17 | 34.15 | 5.57 |
| VNMCE+T5 [14] | / | / | / | / | 26.01 | / | / | / | 74.07 | 39.10 | 26.01 | 41.62 | 35.49 | 5.78 |
| ED [17] | 21.53 | 22.22 | 23.20 | 24.78 | 25.75 | 26.29 | 27.66 | 29.74 | 75.46 | 42.96 | 25.75 | 41.77 | 42.59 | 10.97 |
| FN [11] | 23.47 | 24.07 | 24.68 | 25.66 | 26.27 | 26.87 | 27.88 | 28.96 | 74.84 | 40.87 | 26.27 | 35.30 | 37.77 | 3.64 |
| EL [22] | 24.68 | 25.68 | 26.41 | 27.35 | 28.56 | 30.27 | 31.50 | 33.55 | 75.66 | 43.72 | 28.56 | 38.70 | 40.32 | 8.62 |
| RL [29] | 25.95 | 26.49 | 27.15 | 28.48 | 29.61 | 30.81 | 31.86 | 32.84 | 76.79 | 44.53 | 29.61 | 40.80 | 40.87 | 10.64 |
| RU-LSTM [15] | 25.44 | 26.89 | 28.32 | 29.42 | 30.83 | 32.00 | 33.31 | 34.47 | / | / | 30.83 | / | / | / |
| ActionBanks [36] | / | / | / | / | 28.60 | / | / | / | / | / | 28.50 | / | / | / |
| SF-RULSTM [31] | 26.78 | / | 29.25 | / | 32.05 | / | 34.34 | / | / | / | 32.05 | / | / | / |
| SRL [32] | 25.82 | 27.21 | 28.52 | 29.81 | 31.68 | 33.11 | 34.75 | 36.89 | 78.90 | 47.65 | 31.68 | **42.83** | 47.64 | 13.24 |
| AVT [18] | / | / | / | / | 28.10 | / | / | / | / | / | 28.10 | / | / | / |
| DCR [46] | / | / | / | / | 30.80 | / | / | / | / | / | 30.80 | / | / | / |
| Ours(VS-TransGRU:ES) | 26.84 | 28.01 | 29.48 | 30.41 | 32.36 | 33.57 | 34.91 | 35.82 | 78.08 | 48.19 | 32.36 | 42.04 | 47.95 | 13.76 |
| Ours(VS-TransGRU:GTS) | **35.92** | **37.47** | **38.05** | **38.29** | **40.06** | **40.12** | **40.99** | **42.06** | **79.65** | **58.57** | **40.06** | 42.74 | **59.40** | **19.20** |

## 4 EXPERIMENTS

### 4.1 Datasets and Metrics

**EPIC-Kitchens Dataset.** EPIC-Kitchens [10] is a large-scale first-person dataset collected from 32 participants performing cooking activities in 32 different kitchens, with a total duration of 55 hours. This dataset contains 2513 action categories, 125 verb categories, and 352 noun categories. Since the test set labels are not available, following [15], we randomly select 232 and 40 videos from the publicly available training set for training and validation, respectively. Ultimately, out of the 28,472 usable video clips, 23,493 are used for training and 4,979 for validation.

**EGTEA Gaze+ Dataset.** EGTEA Gaze+ [26] is also a first-person dataset. The dataset consists of 10,325 annotated video clips, with 19 verb categories, 51 noun categories, and 106 action categories. The dataset provides three splits for training/testing, and we report the average performance across the three splits.

**Metrics.** On the EPIC-Kitchens dataset, we evaluate the performance of our proposed method using Top-5 accuracy and Mean Top-5 Recall. For the EGTEA Gaze+ dataset, we use Top-5 accuracy to evaluate the results at different anticipation times.

### 4.2 Implementation Details

As described in Section 3.1, following [15], we take 14 time-steps of inputs before the target action starts, with an interval of 0.25s between every two time-steps. The anticipation steps are set from 1 to 8, resulting in 8 different anticipation times: 0.25s, 0.5s, 0.75s, 1.0s, 1.25s, 1.5s, 1.75s, and 2.0s. To fairly compare with other methods, we also use the pre-trained TSN model provided by [15] as the visual feature extractor $\varphi$. To obtain the ground truth semantic features, we use the pre-trained Phrase-Bert model [44] as the language model $\psi$ in Figure 3. The language model $\psi$ and the visual feature extractor $\varphi$ are fixed during training. [15] provides features including RGB, optical flow, and object features on the EPIC-Kitchens dataset, as well as RGB and Optical Flow features on the EGTEA Gaze+ dataset. For simplicity, we only use RGB features on the EPIC-Kitchens dataset. To be consistent with other methods, we use RGB and optical flow features on the EGTEA Gaze+ dataset. The number of Transformer blocks $M$ is set to 1. The dimension of the GRU hidden state is consistent with the dimension of the semantic enhanced visual features. When using estimated semantic features, the loss function weights $a, b, c$ on the EPIC-Kitchens dataset are set to 2.1, 1.0, and 1.0, respectively, while 2.9, 1.0, and 1.1 on the EGTEA Gaze+ dataset. We use the SGD optimizer to update the parameters, with a learning rate of 0.01 and a momentum of 0.9. The training batch size is 128. After setting the hyperparameters, we train for 100 epochs. Then, we select the best epoch for subsequent testing of other anticipation times based on the Top-5 accuracy at an anticipation time of 1s as the validation metric. More details can be found in our supplementary material.

### 4.3 Comparison with State-of-the-Arts

**Results on EPIC-Kitchens.** We compare our approach on this dataset with state-of-the-art methods, as shown in Table 1. In addition to comparing the Top-5 action accuracy for 8 different anticipation times (i.e., 0.25s-2.0s), we also compare the Top-5 accuracy and Mean Top-5 Recall (class-agnostic and class-aware) for actions, verbs, and nouns at the anticipation time of 1s. Bolded and underlined items indicate the highest and the second-best results, respectively. The results of these methods on RGB features are reported. As shown in Figure 4, our method VS-TransGRU:GTS significantly outperforms other methods when using ground truth semantic features. Specifically, for the Top-5 accuracy of 8 anticipation times, our method achieves an average improvement of 9.04% and 8.15% compared to the previous state-of-the-art methods RU-LSTM [15] and SRL [32], respectively. It is worth noting that SRL only improves by 0.89% on average compared to RU-LSTM. For the Top-5 accuracy and Mean Top-5 Recall evaluation metrics of verbs, nouns, and actions, we achieve an average improvement of 6.28% compared to the previous state-of-the-art methods. Even when using estimated semantic features, our method still performs excellently. For example, among the 14 evaluation metric items, VS-TransGRU:ES achieves the highest performance in 11 items. This further validates the effectiveness of our proposed method.

Table 2: Anticipation results of different methods on EGTEA Gaze+ dataset.

| Methods | Top-5 Action Accuracy(%) @ Different Anticipation Times | | | | | | | |
|---|---|---|---|---|---|---|---|---|
| | 2.00 | 1.75 | 1.50 | 1.25 | 1.00 | 0.75 | 0.50 | 0.25 |
| DMR [42] | / | / | / | / | 55.70 | / | / | / |
| ATSN [10] | / | / | / | / | 40.53 | / | / | / |
| MCE [14] | / | / | / | / | 56.29 | / | / | / |
| ED [17] | 45.03 | 46.22 | 46.86 | 48.36 | 50.22 | 51.86 | 49.99 | 49.17 |
| FN [11] | 54.06 | 54.94 | 56.75 | 58.34 | 60.12 | 62.03 | 63.96 | 66.45 |
| RL [29] | 55.18 | 56.31 | 58.22 | 60.35 | 62.56 | 64.65 | 67.35 | 70.42 |
| EL [22] | 55.62 | 57.56 | 59.77 | 61.58 | 64.62 | 66.89 | 69.60 | 72.38 |
| RU-LSTM [15] | 56.82 | 59.13 | 61.42 | 63.53 | 66.40 | 68.41 | 71.84 | 74.28 |
| LAI [45] | / | / | / | / | 66.71 | 68.54 | 72.32 | 74.59 |
| KDLM [7] | 59.99 | 62.02 | 63.95 | 66.47 | 68.74 | 72.16 | 75.21 | 78.11 |
| SF-RULSTM [31] | 57.48 | / | 61.37 | / | 67.60 | / | 72.22 | / |
| SRL [32] | 59.69 | 61.79 | 64.93 | 66.45 | 70.67 | 73.49 | 78.02 | 82.61 |
| MGRKD [21] | 60.86 | 63.43 | 65.24 | 67.66 | 70.86 | 74.32 | 77.49 | 79.61 |
| HRO [27] | 60.12 | 62.32 | 65.53 | 67.18 | 71.46 | 74.05 | 79.24 | **83.92** |
| Ours(VS-TransGRU:ES) | 61.01 | 63.59 | 65.80 | 67.97 | 71.71 | 74.51 | 79.30 | 82.38 |
| Ours(VS-TransGRU:GTS) | **62.57** | **64.23** | **66.34** | **68.40** | **72.68** | **75.11** | **79.89** | 82.77 |

**Results on EGTEA Gaze+.** We report the Top-5 accuracy for 8 different anticipation times on this dataset, as shown in Table 2. It can be seen that VS-TransGRU:GTS and VS-TransGRU:ES achieve the highest results in 7 anticipation times. VS-TransGRU:GTS improves the average results across all anticipation times by 1.02% compared to the second-best method HRO [27]. When using model-estimated semantic features, the average performance of our method (i.e., VS-TransGRU:ES) is 5.55%, 1.07%, and 0.85% higher than state-of-the-art methods RU-LSTM [15], SRL [32] and MGRKD [21], respectively. When the anticipation time is 0.25 seconds, our method is slightly inferior to HRO. The One-shot Transferring mechanism of HRO takes advantage of the target moment and its subsequent features, while we do not use this information. We believe this is why HRO performs better at the 0.25s anticipation time. The performance improvement brought by introducing semantic features on the EGTEA Gaze+ dataset is less significant than that on the EPIC-Kitchens dataset. This is because the EGTEA dataset has only 106 action categories, while the EPIC-Kitchens dataset has 2513 categories. We infer that this makes the performance improvement relatively smaller, consistent with the views of [15].

### 4.4 Ablation Studies

To further validate the effectiveness of our proposed method, we conduct ablation studies on the EPIC-Kitchens dataset. The experiment results are based on RGB features.

**Model Architecture.** In action anticipation tasks, the mainstream models are based on single RNN. To validate the effectiveness of our designed Transformer-GRU structure, we compare it with other structures, and the experimental results are shown in Table 3. We show the Top-5 accuracy of seven network structures at an anticipation time of 1s. We can see significant differences in anticipation performance among different structures. When using GRU or LSTM as the encoder and decoder, the result is 30.83%, consistent with the results of [15]. When using the original Transformer's encoder and decoder structure, we tried two decoding ways: basic serial decoding and parallel decoding [19]. It can be seen that parallel decoding improves the performance compared to serial decoding, but both are inferior to RNN. We believe this is due to the Transformer structure's inability to flexibly cope with variable anticipation times. Next, we verify the effect of the hybrid structure, i.e., Transformer Encoder + GRU and ViT Encoder [12] + GRU (where ViT Encoder adopts learnable position encoding and Pre-Norm, while Transformer Encoder adopts fixed position encoding and Post-Norm). In our task, the original Transformer Encoder's result is 0.33% higher than ViT Encoder's and 0.87% higher than the RNN-based method. The Transformer Encoder + GRU structure fully utilizes the powerful sequence modeling capabilities of the Transformer and the flexible decoding capabilities of the GRU, which is why it achieves the best results. The performance of our proposed Transformer Encoder + GRU structure at the anticipation time of 1s has already surpassed all methods in Table 1 except SF-RULSTM [31]. It is worth noting that SF-RULSTM uses two complementary visual inputs with different frame rates, while we only use one. This again demonstrates the effectiveness of our proposed network structure. For comparison with state-of-the-art and subsequent ablation experiments, we use this structure by default. In addition, we also tried using a single Transformer Encoder for both encoding and decoding, but the performance was unsatisfactory, with only 30.09%. This further validates the superiority of the encoder-decoder structure, which allows the encoder to focus on modeling the observed data while enabling the decoder to focus on anticipating the future. In contrast, having a single model complete both encoding and decoding is more difficult.

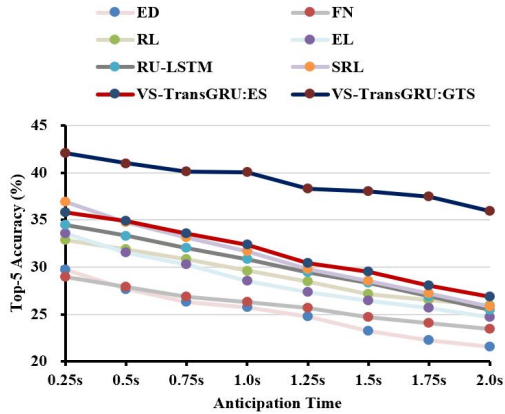

Figure 4: Comparison with state-of-the-art methods at 8 anticipation times on EPIC-Kitchens dataset.

Table 3: The effect of different architectures.

| Encoder | Decoder | Top-5 Act. Acc.(%) @ 1s |
|---|---|---|
| GRU | GRU | 30.83 |
| LSTM | LSTM | 30.83 |
| Transformer Encoder | Transformer Decoder | 30.55 |
| Transformer Encoder | Transformer Decoder(parallel) | 30.79 |
| Transformer Encoder | GRU | **31.70** |
| ViT Encoder | GRU | 31.37 |
| Transformer Encoder | / | 30.09 |

**Table 4: The improvement by introducing semantic with different fusion strategies.**

| Fusion Strategy | Top-5 Action Acc.(%) @ 1s |
|---|---|
| only RGB | 31.70 |
| only Semantic | 35.13 |
| Concatenation | 38.39 |
| Learnable Weighted Sum | **40.06** |
| MLP Fusion | 37.75 |
| Attention Fusion | 36.28 |

**Table 5: The ablation experimental results of loss function.**

| Tgt Loss | Obs Loss | Cos Loss | MSE Loss | Top-5 Action Acc.(%) @ 1s |
|---|---|---|---|---|
| ✓ | | | | 31.75 |
| ✓ | ✓ | | | 31.99 |
| ✓ | | ✓ | | 31.91 |
| ✓ | | | ✓ | 31.85 |
| ✓ | ✓ | ✓ | | 32.25 |
| ✓ | ✓ | | ✓ | 32.13 |
| ✓ | | ✓ | ✓ | 31.97 |
| ✓ | ✓ | ✓ | ✓ | **32.36** |

**Semantic and Fusion Strategy.** We introduce semantic information to egocentric action anticipation for the first time to break the performance bottleneck. We propose various fusion strategies to eliminate the gap between visual and semantic features, as described in Section 3.4, including concatenation, learnable weighted sum, MLP fusion, and attention fusion. In this part, we report the results of fusing ground truth semantic features with visual features. As shown in Table 4, the first row shows the results without incorporating semantic information, i.e., using only RGB features. The second row shows the results using only semantic features. Semantic features achieve better results than RGB, which again demonstrates the validity of the proposed semantic features. It can be seen that combining visual and semantic features can significantly improve the model's anticipation performance, with a maximum improvement of 8.36%. The significant improvement also indicates there is complementary information between visual and semantic features. In terms of fusion strategies, the simple concatenation and weighted sum methods outperform the MLP and attention fusion methods. This might be because the complex fusion methods introduce interference to the model's data modeling. The learnable weighted sum fusion method allows the model to learn suitable weights for different modalities, providing higher flexibility, and it achieves the best results in our experiments. Therefore, we choose it as the fusion method for our framework.

**Loss Function.** When fusing estimated semantic features with visual features, in order to better supervise the model training, we not only use the anticipation loss for the target action but also add the observation loss for the observed action, as well as the cosine similarity and mean squared error loss for the estimated semantic features. We report the ablation results of the loss function based on the learnable weighted sum fusion strategy. As shown in Table 5, adding any of the three loss items will improve the anticipation performance, with the most significant improvement coming from adding the observation loss. Then, we find that combining two losses works better than a single loss, indicating complementarity between different losses. When all the three additional loss items are used, the anticipation performance reaches its optimal level.

More additional ablation studies can be found in our supplementary material.

## 5 QUALITATIVE RESULTS

Please see our supplementary material for qualitative results.

## 6 CONCLUSION

This paper proposes a novel Transformer-GRU-based framework enhanced by visual-semantic fusion for egocentric action anticipation. For the first time, we introduce semantic features to assist action anticipation, which can effectively improve the model's anticipation performance. Semantic features can be obtained either from category labels through language models or estimated based on visual information. Unlike previous methods solely relying on a single recurrent neural network structure, we designed a hybrid model based on Transformer and GRU with more robust modeling and anticipation capabilities. Moreover, we propose various fusion strategies to eliminate the gap between visual and semantic features. The fused features are then fed into the Transformer-GRU structured model to generate results for future actions. Extensive experiments on two large-scale datasets demonstrate the effectiveness of our proposed method.

# Supplementary Material for "VS-TransGRU: A Novel Transformer-GRU-based Framework Enhanced by Visual-Semantic Fusion for Egocentric Action Anticipation"

## 1 MORE IMPLEMENTATION DETAILS

After inputting the semantic-enhanced visual feature sequence, to obtain the pooled features, we take the mean value of the Transformer's outputs in the sequence length dimension on the EPIC-Kitchens dataset and take the value of the Class Token on the EGTEA Gaze+ dataset as the pooled features. We use RGB and optical flow modalities on the EGTEA Gaze+ dataset. On this dataset, we train models separately for each modality and then train learnable weight parameters to fuse the anticipation results of the two modalities.

## 2 ADDITIONAL ABLATION STUDIES

We report the results of this section using RGB features and the Transformer + GRU model on the EPIC-Kitchens dataset unless otherwise specified.

**The Number of Transformer Layers.** We use the Transformer to encode the observed data. One important hyperparameter in Transformer is the number of Transformer layers, denoted as $M$. Figure 1 shows the anticipation results for different $M$ (from 1 to 6). It can be seen that as $M$ increases, the anticipation performance decreases. This is because increasing the number of layers also increases the number of parameters, leading to an overfitting phenomenon.

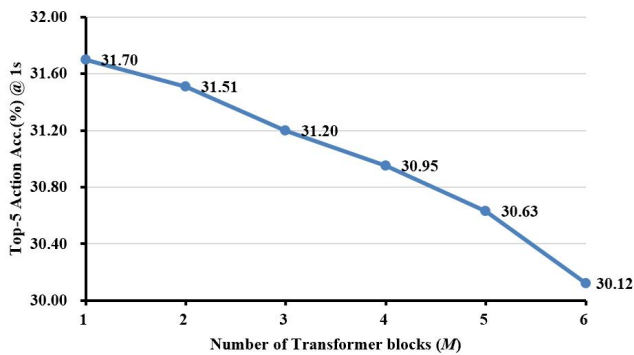

Figure 1: The results of varying the number of Transformer blocks.

**Label Smoothing.** The common practice is using the cross-entropy loss to anticipate actions at the target time. We use a cross-entropy loss function with label smoothing to prevent the model from being overly confident in its anticipation results to improve anticipation performance. As shown in Table 1, label smoothing does bring performance improvements.

**Feature Dimension Projection.** Since the semantic features we use have a different dimension from the visual features, when the two need to be fused while maintaining the same dimension, we can either project the semantic features to the dimension of the visual features or project the visual features to the dimension of the semantic features. We conducted experiments for both methods, and the results are shown in Table 2. We report the results in Table 2 based on the ground truth semantic features and learnable weighted sum fusion strategy. As can be seen, the method of projecting semantic features performs better, so we adopt this approach in our experiments.

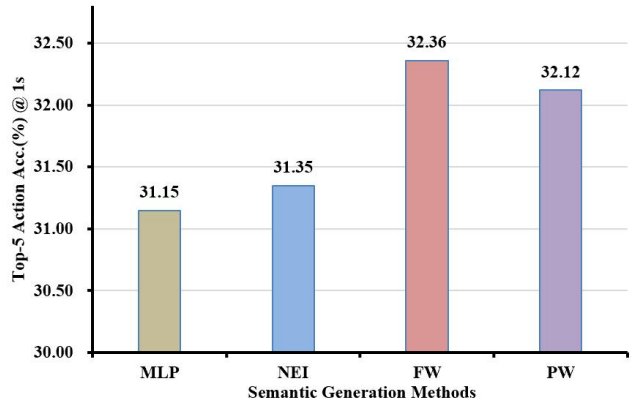

Figure 2: The results of different semantic generation methods.

Table 1: The ablation experimental results of label smoothing.

| Label Smoothing | Top-5 Action Acc.(%) @ 1s |
| --- | --- |
| without | 31.55 |
| with | **31.70** |

**Semantic Generation Methods Based on Visual.** As for how to enable the model to generate semantic features based on visual information, we consider the following methods:

1. Add an MLP module to generate semantic features based on visual features directly and produce cosine similarity and mean squared error losses between the generated and ground truth semantic features.

2. We convert the model's classification score vector for the current action into a numerical encoding of the category. Then, we use this numerical encoding to index the ground truth semantic feature sequence corresponding to the category labels, obtaining the estimated semantic features. For example, consider a situation where the model's recognition result for the current action assigns the highest value to category 2. This implies that, according to the model, the current action is most likely to belong to category 2.

| | Observations | | | Future Target Actions | Anticipation Results |
|---|---|---|---|---|---|

Figure 3: Visualization of the anticipation results of our method and baseline. The ground truth categories, correct and incorrect anticipations are marked in blue, green, and red, respectively.

Table 2: The ablation experimental results of feature dimension projection.

| Projection Strategy | Top-5 Action Acc.(%) @ 1s |
|---|---|
| Project RGB | 39.13 |
| Project Semantic | **40.06** |

Consequently, the ground truth semantic features corresponding to category 2 are taken as the model's estimated semantic features. We call this method "Numerical Encoding Index (NEI)".

3. As described in the main text, we use the values of the model's recognition scores for the current action after the Softmax activation function as weights and multiply them with the ground truth semantic feature matrix to obtain the estimated semantic features. We call this method "Full Weights (FW)".

4. Combining methods 2 and 3, in the model's classification score vector for the current action, we apply the Softmax activation function to the highest $K$ category classification values to obtain the weights of the $K$ categories. Then, by setting the weights of other categories to 0, we expand the $K$-dimensional vector into an $N$-dimensional vector, where $N$ represents the number of action categories and $K < N$. Finally, the $N$-dimensional vector is multiplied by the ground truth semantic feature matrix to obtain the estimated semantic features. We call this method "Partial Weights (PW)". In the experiments, we set the values of $K$ to 10, 50, 100, 200, 500, and 800. The experimental results show that the PW achieves its best performance of 32.12% when $K$ is 500. Therefore, we use 32.12% as the result of the PW reported in Figure 2.

As shown in Figure 2, MLP and NEI achieve unsatisfactory results. For MLP, we believe it is tough to directly generate higher-order semantic features of multiple categories based on visual features. It has certain shortcomings for NEI: when the observed action's recognition result is incorrect, the obtained semantic features will be incorrect. This is the reason why NEI's performance is relatively poor. In contrast, FW and PW alleviate this problem better, with FW achieving the best and PW achieving suboptimal results. There is a slight difference in anticipation performance between the FW and PW methods, indicating that using the anticipation results of all categories as weights is better than using only some categories.

## 3 QUALITATIVE RESULTS

To understand our proposed method better, we visualize its anticipation results on the EPIC-Kitchens dataset. At the same time, we use the GRU-GRU (i.e., two GRUs serving as encoder and decoder, respectively) model based on visual information as the baseline and visualize its anticipation results, too. Their visualization results are shown in Figure 3. The observation time is 2.75s, and the anticipation time is 1s. We visualize not only correct anticipation results but also incorrect ones. In the first six examples of Figure 3, the baseline generates three incorrect anticipation results, while our method can accurately anticipate all of them. This highlights the superiority of our proposed visual-semantic fusion method and Transformer-GRU architecture, which can effectively understand the observed data and better capture the dependencies between current and future actions. However, our method may also produce incorrect anticipations. For example, in the seventh row of Figure 3, when the observed action is "wash spoon" and there is a pan to be cleaned nearby, both our VS-TransGRU:ES and VS-TransGRU:GTS believe that the future action is "wash pan". Nevertheless, the performer puts down the sponge and stops washing actually, which is unscripted. It is evident that action anticipation on unscripted datasets like EPIC-Kitchens is more challenging than on scripted datasets. In the last row of Figure 3, when the observed action is "open door", only "cup" appears in the following field of view, and our VS-TransGRU:ES and VS-TransGRU:GTS anticipate the subsequent action to be "close door" and "take cup", respectively. Interestingly, the ground truth next action is "take plate" but the "plate" never appears visually. This situation is caused by the angle of the performer's camera. In this case, it is extremely difficult for the model to anticipate "take plate" accurately, and even humans will likely make errors.